\documentclass{article}

\usepackage{amsmath}

\usepackage{arxiv}
\usepackage[utf8]{inputenc} 
\usepackage[T1]{fontenc}    
\usepackage{hyperref}       
\usepackage{url}            
\usepackage{booktabs}       
\usepackage{amsfonts}       
\usepackage{nicefrac}       
\usepackage{microtype}      
\usepackage{lipsum}
\usepackage{graphicx}
\graphicspath{ {./images/} }
\usepackage{algorithm}
\usepackage{algpseudocode}

\usepackage{adjustbox}
\usepackage{multirow}
\usepackage{multicol}
\usepackage{amssymb}
\usepackage{bbm}
\usepackage{float}
\usepackage{subcaption}
  
\title{Autoregressive based Drift Detection Method
}

\author{
  Mansour Zoubeirou A Mayaki \\
  Université Côte d’Azur \\
  CNRS, Inria, I3S \\
  Nice France\\
   \And
  Michel Riveill \\
  Université Côte d’Azur \\ CNRS, Inria, I3S \\ Nice France 
}

\begin{document}
\maketitle
\date{}

\begin{abstract}
In the classic machine learning framework, models are trained on historical data and used to predict future values. It is assumed that the data distribution does not change over time (stationarity). However, in real-world scenarios, the data generation process changes over time and the model has to adapt to the new incoming data. This phenomenon is known as concept drift and leads to a decrease in the predictive model's performance. In this study, we propose a new concept drift detection method based on autoregressive models called \textbf{ADDM}. This method can be integrated into any machine learning algorithm from deep neural networks to simple linear regression model. Our results show that this new concept drift detection method outperforms the state-of-the-art drift detection methods, both on synthetic data sets and real-world data sets. Our approach is theoretically guaranteed as well as empirical and effective for the detection of various concept drifts. In addition to the drift detector, we proposed a new method of concept drift adaptation based on the severity of the drift.
\end{abstract}


\section{Introduction}
Thanks to progress in the field of big data and data analysis, machine learning models and more particularly those based on  deep neural networks (Deep Learning) are nowadays experiencing phenomenal success. 
Since the beginning of the 2010s, neural networks have been developing at high speed and the fields of application are multiplying in all business sectors. In the machine learning framework, models are trained on historical data and used to predict future values. In this framework, we assume that future incoming data streams are stationary, i.e., the data generating process does not change over time. However, this assumption does not hold in most real-world applications  \cite{baier2021detecting}. For example the statistical properties of a streaming data can change over time due to seasonality or random events. This phenomenon is known in the machine learning community as concept drift. In the presence of concept drift, the model's predictions become less accurate over time. 

Machine learning models should therefore take in account concept drift and update their weights at the right time. Detecting a concept drift is one of the main challenges when learning with streaming data because of the high speed and their large size sets which are not able to fit in the main memory  \cite{kadam2019survey}. To deal with concept drift, many algorithms and methods (ADWIN, DDM, KSWIN, PageHinkley) have been proposed in the literature. Most of these algorithms detect concept drifts by tracking the changes in the model's error rate or using a distance function to measure the dissimilarity of the input data distribution between some timestamps. These methods are very sensitive to changes leading to large numbers of detected drifts and false alarms  \cite{baier2021detecting}. Moreover, most of these algorithms require full and immediate access to ground-truth labels which is an unrealistic assumption in  most  real-world  applications.

To accurately detect concept drifts in stream data,  we propose to integrate an autogressive time series model inside the machine learning loop by considering the model's error as a time series. We used a self-exciting threshold auto-regressive (SETAR) model  \cite{tong2009threshold} as the base autoregressive model. SETAR is a nonlinear time series model and a special case of regime switching models in which different models apply to different intervals of values of some key features.
Our method has two components: a machine learning model for the learning task and a SETAR model that detects the changes in the learning model's error rate distribution.
We call the new concept drift detection method \textbf{ADDM}. This approach can be used with any type of predictive model (Logistic regression, random forest etc.).
Our results show that the new concept drift detection method outperforms all state-of-the-art methods on six (6) synthetic data sets and five (5) real-world data sets. ADDM is more accurate and has a very low false alarms rate. A low false alarms rate is very important in real-world application because retraining a machine model is time-consuming and resource-intensive. 
This method also has some theoretically guarantees as the parameters of the change detection component (SETAR model) are estimated using ordinary least squared (OLS) \cite{tsay1989testing}. Another advantage of our method is that we can construct confidence intervals for the detected drift points using statistical inference and subsampling \cite{gonzalo2005subsampling}. In addition to the drift detector, we proposed a new method of concept drift adaptation based on the severity of the drift. The main idea is to aggregate the old and the new models using an estimate of the dissimilarity between the old concept and the new one as weights. The higher is the severity, the less relevant is the old model.

The rest of the paper is outlined as follows. The \textbf{Related Works} section discusses the notions related to concept drift 
and the other studies (or articles) related to concept drifts detection methods/algorithms. The third section is dedicated to the theoretical definition of  the self-exciting threshold auto-regressive model and the description of ADDM concept drift detection method. In section \textbf{Experimental Data sets}, we describe the data sets used for our experiments. The \textbf{hyperparameters optimization} section describes in detail the models architecture, the performance metrics and the drift detection algorithms hyper-parameters optimization. The results are presented and discussed in section \textbf{Results and discussions}.
\section{Related Works}
In statistical (or machine) learning domain, concept drift occurs when the statistical properties of the targeted variable $y$ varies arbitrary over time due to a change in the input data X distribution. Concept drifts can be categorised in three groups according to their sources  \cite{lu2018learning}. The first type of concept drift called virtual drift occurs when the data distribution changes but does not affect the decision boundaries: $P_{t}(X) \neq P_{t+1}(X)$ while $P_{t}(y|X)= P_{t+1}(y|X)$. Virtual concept drift is not well studied in machine learning community because it does not affect the model's outputs. The second source of concept drift called actual drift happens when the drift changes the target variable. Thus the a posterior probability of the data changes in time while it distribution remains unchanged: $P_{t}(y|X) \neq P_{t+1}(y|X)$, $P_{t}(X)= P_{t+1}(X)$. The last type of concept drift result from the mixture of the two first sources: $P_{t}(y|X) \neq P_{t+1}(y|X)$ and  $P_{t}(X) \neq P_{t+1}(X)$.
In practice it is very difficult to separate these sources of concept drifts when learning with stream data. The drift detection algorithms just try to detect the changes that affect the model's output without focusing on their sources.
\subsection{Concept Drift Understanding}
Concept drift understanding answers three main questions: \textbf{When} did the drift occur, \textbf{How} severe is the change and \textbf{Where} are the drifts regions  \cite{lu2018learning}.
The \textbf{when} refers to the fact that any concept drift detection algorithms should be able to detect the timestamps where the data distribution changes significantly. 
Recalling the definition of concept drift, when a drift occurs at time $t$, an alarm signal is triggered and it also indicates that the learning system should adapt to a new concept. Another important question is 
\textbf{how} much did the data distribution change at the drift points (severity of the drifts). The  severity  of  concept  drift  quantifies the  dissimilarity  between  the  new  concept and  the  previous  concept . The severity is defined as $\Delta=\delta(P_{t}(X,y), P_{t+1}(X,y))$ where $\delta$ is a function that measures the discrepancy of two data distributions when there is a drift at timestamp $t$  \cite{lu2018learning}. The greater the value of $\Delta$, the larger the severity of the concept drift. The severity gives an idea of how the learners should adapt to the new concept. If the $\Delta$ is low we may just need to update the learners without changing many parameters. In contrast if the drift is severe, we may need to retrain a whole new model. 
The last question is to identify  \textbf{where} the drift regions (new concepts) are located. The drift regions of concept drift are the sub-regions where the new concept and the previous concept are located. These sub-regions are identified by finding parts of the features space where  $P_{t}(X,y)$ and $ P_{t+1}(X,y)$ are statistically different. In ensemble learning scenarios, detecting concept drift regions can help predicting instances in stable regions.  Moreover, when learning with an artificial neural network we can use the knowledge of the concept drift regions to put weights on the features. 

Like all the state-of-the-art concept drift detection algorithms, ADDM responds to the first question. It detects with high precision the drift points by monitoring the model's error rate. Contrary to the state-of-the-art algorithms, ADDM can compute confidence intervals of each detected drift using statistical hypothesis testing and subsampling methods. 

\subsection{Concept Drift Detection Methods in the Literature}
Many algorithms and methods have been proposed in the literature to detect concept drifts in stream data. These methods can be classified into  three  categories  in  terms of the test statistics they apply \cite{lu2018learning}. The first category called \textbf{error  rate-based  drift  detection  algorithms} refers to all the methods that track changes in the online error rate of base models. The algorithms trigger an alarm when there is a statistically significant increase or decrease of the error rate at some timestamps. Our \textbf{ADDM} method belongs to this category. The second category is \textbf{data distribution-based drift detection} algorithms which use  a distance  function or a metric  to  quantify  the dissimilarity between the distribution of data before and after the suspected drift timestamp  \cite{lu2018learning}. These algorithm detect drifts directly from the input data and try to detect the time and the location of the drifts. The last category called \textbf{multiple  hypothesis  test  drift  detection} methods use multiple hypothesis tests to detect concept drift. The most popular state-of-the-art drift detection algorithms (ADWIN, DDM, KSWIN, PageHinkley) are error  rate-based. The \textbf{ADaptive WINdowing (ADWIN)}  \cite{bifet2007learning} is an adaptive sliding window algorithm for  concept drift detection in stream data. \textbf{ADWIN} require the user to specify a sensitivity hyperparameter $\alpha \in (0,1)$ which allows the algorithm to adjust to the input data. A drift  is  detected when two sub-windows of a recent window of observations exhibit an absolute difference in means larger than $\alpha$. 
The \textbf{Drift Detection Method} (DDM) is a concept drift detection method based on the PAC learning model premise  \cite{gama2004learning}. If the algorithm detects an increase in the error rate higher than a calculated threshold, an alarm is triggered, either change is detected or the algorithm will warn the user that change may occur in the near future. 
The \textbf{Page-Hinkley} (PH) concept drift detector detects changes by computing the observed values and their mean up to the current moment  \cite{page1954continuous}. The algorithm detects a concept drift if the observed mean at some instant is greater than a threshold value $\lambda$. 
The \textbf{Kolmogorov-Smirnov Windowing} (KSWIN) concept drift detection method is based on the Kolmogorov-Smirnov (KS) statistical test  \cite{raab2020reactive}. 
Other versions of these algorithms have been proposed by other authors:  Learning  with  Local  Drift  Detection  (LLDD)   \cite{gama2006learning}, Early  Drift  Detection  Method  (EDDM) \cite{baena2006early},  Heoffding’s inequality  based  Drift  Detection  Method  (HDDM) \cite{frias2014online}, Dynamic  Extreme  Learning  Machine  (DELM)  \cite{xu2017dynamic}.

Baier et~al.  \cite{baier2021detecting} used neural network uncertainty instead of the model's error rate to detect concept drift. The authors proposed to use the \textbf{Monte Carlo Dropout} technique to capture the model's uncertainty. A drift is detected when the model's uncertainty increases or decreases significantly at some timestamp(s). Their algorithm called Uncertainty Drift Detection (UDD) is based on \textbf{ADWIN} algorithm. The main advantage of UDD method is that in contrast of the majority of error based drift detection algorithms, it does not require full and immediate access to ground-truth labels which is an unrealistic assumption in  most  real-world  use  cases  \cite{baier2021detecting}.

Yan et~al. \cite{yan2020accurate} proposed an algorithm based on Hoeffding’s inequality to monitor the error rate and detect concept drift. The main idea of their method is to use Hoeffding’s concentration inequality to examine the consistency of the predictive error. 
This algorithm relies on the theorem that if the data distribution is stationary if the difference between the predictive error at time $t$ denoted  $p_{t}$ as and its lower bound $p_{bayes}$ goes to 0 when  the  number  of  training  instances  increases  \cite{gama2013evaluating}. The authors used Hoeffding’s Inequality to estimate the desired upper bound above which the  error  rate  is considered highly unstable. If the predictive error difference $\Delta_{t}=p_{t}-p_{bayes}$ is  very  high after  learning  a  large  enough  number  of  instances,  it  means that  their is a concept drift and the data distribution has changed. According to there results, their method gives better results than the state-of-the-art methods.
Wang et~al. \cite{6706768} proposed a new concept drift detection method for class imbalanced problem called DDM-OCI. The new method is inspired from the DDM \cite{gama2004learning}, instead of monitoring the model's error rate, the authors used the recall of the minority class to detect changes in the data distribution. Their results show that, DDM-OCI responds to new concepts faster than the model applying DDM. Greco and Tania \cite{9679880} proposed a real-time unsupervised per-label drift detection methodology based on embedding
distribution distances in deep learning models . Their method exploits the inner representations assigned by a deep learning model to new unseen data to detect drifts.

\section{Our approach for learning Under Concept Drift}
\subsection{Self-exciting Threshold Autoregressive (SETAR) Models}
\label{setarSec}
The self-exciting threshold autoregressive model is a nonlinear time series model proposed by Tong in 1978  \cite{tsay1989testing}. This model is special case of regime switching models in which different models apply to different intervals of values of some key variable. This model has certain properties such as limit cycles, amplitude dependent frequencies, and jump phenomena that can not be captured by classic linear time series models  \cite{tong2009threshold}. 
Formally, a  SETAR model with k regimes can be written mathematically as follows  \cite{tsay1989testing}:
\begin{equation}
\label{eq_setar}
    Y_{t}= \phi_{0}^{(i)}+ \phi_{1}^{(i)}Y_{t-1}+\dots +\phi_{p_{i}}^{(i)}Y_{t-p}+\epsilon_{t}^{i}
\end{equation}
\begin{equation}
    r_{i-1}\leq Y_{t-d} <r_{i}
\end{equation}

Where $i=1\dots k$, $1<d\leq \max(p_{i})$ a positive integer, $Y_{t}$ is a time series and $Y_{t-d}$ the threshold variable. The thresholds values are $-\infty < r_{0}< r_{1}<\dots< r_{k}< + \infty$; for each regime $i$, the error term ${\epsilon_{t}^{i}}$ is a sequence of martingale differences satisfying:
\begin{equation*}
    E(\epsilon_{t}^{i}|F_{t-1})=0, \hspace{0.3cm} sup_{t}=E(|\epsilon_{t}^{i}|^{\delta}|F_{t-1}|) < \infty 
\end{equation*}
\begin{equation*}
 \hspace{0.3cm} a.s. \hspace{0.3cm} for \hspace{0.3cm} some \hspace{0.3cm} \delta>2
\end{equation*}
With $F_{t-1}$ a $\sigma$ field generated by $\{\epsilon_{t-j}^{i}|j=1,\dots;i=1,\dots,k\}$. Such a process partitions the one-dimensional Euclidean space into k regimes and follows a linear auto-regressive model in each regime  \cite{tsay1989testing}. A two regime SETAR model can be written as follows:

\begin{equation}
\label{setar}
    Y_{t} =( \phi_{0}+ \phi_{1}Y_{t-1}+\dots +\phi_{p}Y_{t-p})\mathbbm{1}_{Y_{t-d}\leq r} 
     +( \beta_{0}+ \beta_{1}Y_{t-1}+\dots+\beta_{p}Y_{t-p})\mathbbm{1}_{Y_{t-d}>r}+\epsilon_{t}
\end{equation}

Where $p$ denotes the autoregressive level , $Y_{t-d}$ the threshold variable and $r$ the threshold parameter. In principle, we would like $\epsilon_{t}$ to be conditionally heteroskedastic, but for formal theory, we assume that $\epsilon_{t}$ is $iid$ $(0,\sigma^{2})$.  This model is called self-exciting because the threshold variable is a function of the past values of the endogenous variable $Y_{t}$.
Since the SETAR model is a locally linear model, ordinary least squares (OLS) techniques can be used to estimate its parameters \cite{hansen1996inference}. Under the assumption that the error $\epsilon_{t}$ is $iid$ $N(0,\sigma^{2})$, OLS is equivalent to maximum likelihood estimation.  
The model in equation~\ref{eq_setar} can be rewritten as follows:
\begin{equation}
    Y_{t}=\Phi^{'}X_{t}(r)+\epsilon_{t}
\end{equation}

\begin{equation*}
    X_{t}=(1,Y_{t-1},\dots,Y_{t-p})^{'}
\end{equation*}

\begin{equation*}
    X_{t}(r)=( X_{t}\mathbbm{1}_{Y_{t-d}\leq r}, X_{t} \mathbbm{1}_{Y_{t-d}> r})
\end{equation*}

Where $\Phi=(\phi,\beta)$, $\phi=(\phi_{0},\phi_{1},\dots, \phi_{p})$ and $(\beta=\beta_{0},\beta_{1},\dots, \beta_{p})$.

The parameters of interest are $\Phi$ and $r$. For a given threshold value $r$, the OLS estimate of $\Phi$ is  \cite{hansen1996inference}

\begin{equation}
\label{eq3}
    \Hat{\Phi(r)}=\left( \sum_{t=1}^{T}X_{t}(r)^{'}X_{t}(r) \right)^{'}\left( \sum_{t=1}^{T}X_{t}(r)^{'}Y_{t} \right)
\end{equation}
With estimated residuals $\hat{\epsilon_{t}}(r)=Y_{t}-X_{t}(r)^{'}\Hat{\Phi(r)}$ and their estimated variance 
\begin{equation}
\label{threshold}
    \hat{\sigma_{T}^{2}}(r)=\frac{1}{T}\sum_{t=1}^{T}\hat{\epsilon_{t}}^{2}
\end{equation}
The estimation task is now reduced to finding the threshold values $r$ that minimizes estimated residuals variance $\hat{\sigma_{T}^{2}}(r)$ which  depends exclusively on $r$. The other parameters are then computed by using equation $\eqref{eq3}$. The threshold parameter is computed by minimizing equation $\eqref{threshold}$ as follows:
\begin{equation}
\label{thresh_eq}
    \hat{r}=\arg \min_{r\in R} \hat{\sigma_{T}^{2}}(r)
\end{equation}
Where $R=\{ y_{t-d}, \text{for all} \hspace{0.3cm} d+1\leq t\leq T \}$. $R$  has a finite number of elements $[T-(d+1)]$.

Since the model's parameters are estimated using ordinary least squared (OLS) method, we have some theoretical guarantees of its convergence \cite{tsay1989testing}. We can also construct confidence intervals for the detected drift points by using statistical inference  and sub-sampling\cite{gonzalo2005subsampling}.

\subsection{ADDM for Concept Drift Detection}
Our \textbf{ADDM} method belongs to the category of concept drift detection methods based on error rate monitoring. The model triggers an alarm when there is a statistically significant increase or decrease of the error rate at some timestamps. As proved by Gama et al.\cite{gama2013evaluating}, if the data distribution is stationary, the error rate should converge to its minimum value. Therefore if the error rate becomes very high after predicting a large enough sample, it may indicate that the data distribution has changed and the model is no longer fit to the data.
ADDM has two components: a machine learning model for the prediction task and a SETAR model that detects the changes in the learning model's error rate $Y_{t}$.
When new data instances arrive, they are predicted at the time of arrival with the deep learning model. The prediction errors are then computed and used as target variable $Y_{t}$ for the SETAR model (see Fig.1). Referring to the definition of the SETAR model in equation \eqref{eq_setar}, our model forecasts the error rate  using a linear combination of its past values assuming that the behavior of the error rate changes once it enters a different regime or concept. The threshold values estimated by equation \eqref{thresh_eq} correspond to the concept drift points in the data. We can also construct confidence intervals for each detected drift using statistical inference and sub-sampling \cite{gonzalo2005subsampling}.



\begin{figure}[!htp]
\centering
\includegraphics[width=\columnwidth]{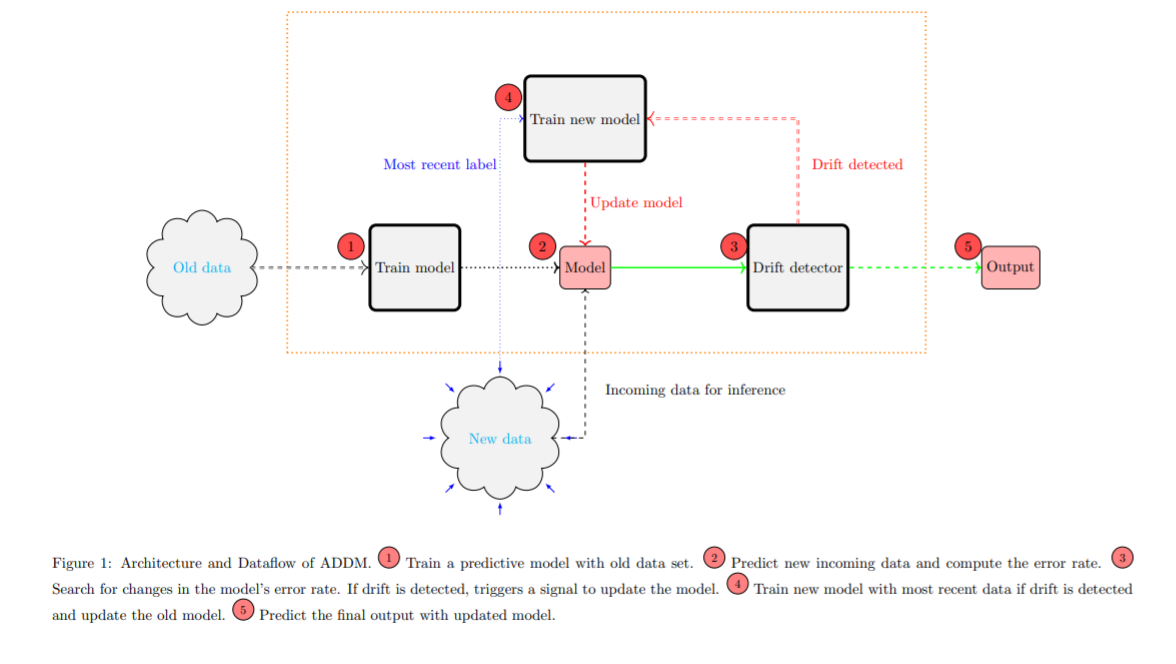}
\caption{Architecture and Dataflow of ADDM. (1) Train a predictive model with old data set. (2) Predict new incoming data and compute the error rate. (3) Search for changes in the model's error rate. If drift is detected, triggers a signal to update the model.
(4) Train new model with most recent data if drift is detected and update the old model. (5) Predict the final output with updated model.}
\label{modelflow}
\end{figure}

\subsection{Model Updating or Concept drift adaptation}
After detecting concept drifts, we need to update the model so that it adapts to the new data distribution. In the literature there are mainly three groups of drift adaptation methods. The first method consist of retraining a whole new model when a concept drift is detected. The second strategy is ensemble method witch consist of aggregating a new model trained on the samples from the new distribution with the old  models. This strategy can  save  significant  effort  to  retrain  a  new model for recurring concepts. Ensemble  methods comprise  a  set  of  base  classifiers  that  may  have  different types  or  different  parameters \cite{lu2018learning}.  The last drift adaptation strategy consists of partially updating the model when  the  underlying  data  distribution  changes. This strategy is  more  efficient  than retraining an entire model when the drift only occurs in well located regions (decision  tree algorithm are suited in this case because trees have the ability to examine and adapt to each sub-region separately). This approach can be difficult to use in case of deep learning models because these models are considered as black boxes and we don't know witch parameters to update.

We propose a new concept drift adaptation method that consist of aggregating the old model with a new model trained on the most recent samples using the severity of the concept drift. The main idea is aggregate the old and new model using an estimate of the dissimilarity (severity of the drift) between the old concept and the new one as weight. The higher is the severity, the less relevant is the old model.
The severity denoted $w_{t}$ is also used as the weight of the new model in the final model. We first compute the third quantile $Q_{3}$ in each regime and compute $w_{t}$ as follows:

\begin{equation}
    w_{t}=\frac{\max(Q_{3}^{0},Q_{3}^{t})}{Q_{3}^{0}+Q_{3}^{t}}
\end{equation}
Where $Q_{3}^{0}$ is the third quantile of the error rate in the old concept and $Q_{3}^{t}$ in the new concept. We used the third quantile $Q_{3}$ just to make sure that we have a good estimate of the error rate in each regime. One can use any other quantile or aggregating metrics (mean, variance etc.). The quantiles are more suited because they are less sensitive to extreme values. The term $\max(Q_{3}^{0},Q_{3}^{t})$ is used to ensure that the model learned under the new concept always gets the highest weight during aggregation.
The final model is defined as follows:
\begin{equation}
    M_{\hat{\phi}}^{t} \leftarrow M_{\hat{\phi}}^{0}\cdot (1-w_{t})+ w_{t}\cdot M_{new}
\end{equation}
Where $M_{\hat{\phi}}^{0}$ is the old model and $M_{new}$ the new model trained with the most recent data. The new model $M_{new}$ can be learned on a subset (or a window) containing the most recent data or on the whole data set. The main advantage of our method is that it takes into account the severity of the drift when updating the model. If the drift is very severe, the new model has a much more important role than the old one and the influence of the old model may fade. Another advantage is that, it's very flexible and can be used with any kind of model. For example when learning with artificial neural networks, we can average the old and the new model parameters using $w_{t}$ or just average their outputs.

\textbf{ADDM} algorithm is defined as follows:
 
 \begin{algorithm}
	\caption{ADDM algorithm} 
	\begin{algorithmic}[1]
	 \State  \textbf{Input:} Training data $D_{tr}$; Validation data $D_{val}$; Data stream $D_{s}$
	 \State Train deep learning model: $M_{\hat{\phi}}^{0}\leftarrow M_{\phi}\cdot fit(D_{tr},D_{val})$
	 \State  Compute validation error: $\hat{\epsilon}_{val}\leftarrow (y_{val}-\hat{y}_{val})^{2}$
	 \State Fixe a time window $w$ 
		\Repeat 
		    \State Receive incoming data instances $x_{t-w}$ 
		    \State Predict values: $\hat{y}_{t-w} \leftarrow M_{\hat{\phi}}^{0}\cdot predict(x_{t-w})$ 
		    \State  Compute error: $\hat{\epsilon}_{t-w}\leftarrow (y_{t-w}-\hat{y}_{t-w})^{2}$
		    
		    \State  Learn Setar model with $\hat{\epsilon}_{t-w} \cup \hat{\epsilon}_{val}$ 
			\If {change is detected}
			    \State Compute drift severity: $w_{t}\leftarrow\frac{\max(Q_{3}^{0},Q_{3}^{t})}{Q_{3}^{0}+Q_{3}^{t}}$
				\State  Get most recent labeled data $D_{recent}$
				\State Train new model $M_{new}\leftarrow M_{\phi}\cdot fit(D_{recent})$
				\State $M_{\hat{\phi}}^{t} \leftarrow M_{\hat{\phi}}^{0}\cdot (1-w_{t})+ w_{t}\cdot M_{new}$
			\EndIf
		\Until{$D_{s}$ ends}
	\end{algorithmic} 
\end{algorithm}

\section{Experimental Data sets}
\label{data}
In order to evaluate our method's capabilities, we compared its performance to those of seven (7) state-of-the-art methods on six (6) synthetic data sets with artificial concept drifts and five (5) real-world data sets. The synthetic data sets were simulated using the python scikit-multiflow package \cite{skmultiflow}.
The \textbf{Friedman} multi-variate regression data set  \cite{friedman1991multivariate} consist of teen features each generated from a uniform distribution from the interval $[0,1]$. 
The Friedman data set is commonly used to test concept drift detection methods . In our experiments, we simulated three different versions of the Friedman data sets with different types of drifts.  The \textbf{Brieman} regression data set is inspired by Baeir et~al.  \cite{baier2021detecting}. The data set contains teen features, simulated from uniform distribution. The \textbf{Mixed} data set was inspired by Gama et~al.  \cite{gama2004learning} and has 6 attributes. Four attributes are relevant for classification: two boolean attributes and two numeric attributes uniformly distributed from 0 to 1  \cite{skmultiflow}. The \textbf{Agrawal} stream generator was first introduced by Agrawal et~al.  \cite{agrawal1993database}. The generator generates a stream data set of nine features, six numeric and three categorical  \cite{skmultiflow} for binary classification task. We generated two data sets from this generator with different types of concept drift.
At the end we have six synthetic data sets among which three regression data sets and three classification data sets. In each data set, artificial concept drifts were introduced by modifying the distribution of some features. 

In addition to these synthetic data sets, we tested our method on five real-world data sets. Note that all the following data sets are publicly available on UCI Machine Learning Repository website  \cite{Dua2019}. The \textbf{Panama electricity}  data set contains historical records of Panama's electricity demand and weather measures from January 2015 until June 2020  \cite{aguilar2021short}. The data set contains historical electricity load, calendar information related to holidays (and school period) and Weather variables, such as temperature, relative humidity, precipitation, and wind speed, from three main cities in Panama  \cite{aguilar2021short}. The goal is to predict the electricity demand using all available features. In this data set,  concept drift is present due to seasonal weather changes which affects the electricity demand. 
The \textbf{Italian air quality} data set contains the responses of a gas multi-sensor devices deployed on the field in the Italian main cities. Hourly responses averages are recorded along with gas concentrations references from a certified analyzer \cite{de2008field}. The data were recorded from March 2004 to February 2005. They recorded some air quality measures such as the hourly averaged concentrations for CO, Non Metanic Hydrocarbons, Benzene, Total Nitrogen Oxides (NOx) and Nitrogen Dioxide (NO2). The goal is to predict the benzene concentration (C6H6(GT)), which is a proxy for air pollution. Like in the electricity data set, concept drift is present due to seasonal weather changes. According to the World Air Quality Index project, the air is very polluted in the Italian main cities between November and February  \cite{aqicn}. The \textbf{NSW} data set contains data from the Australian New South Wales electricity market  \cite{gama2004learning}. In this market, prices are flexible and are affected by demand and supply of the market. The data set contains 45.312 instances and nine features dated from 7 May 1996 to 5 December 1998. 
The goal is to predict if the electricity price goes up or down each 30 minutes. Concept drift is present due to seasonal weather changes which affects the electricity demand and its price. The \textbf{gas sensor array drift} data set contains measurements from 16 chemical sensors utilized in simulations for drift compensation in a discrimination task of 6 gases at various levels of concentrations. The data set was gathered over a period of 36 months in a gas delivery platform facility situated  \cite{vergara2012chemical}. The goal is to achieve good classification performance over time. Concept drifts are present in the data due to sensor aging and external alterations. The \textbf{Beijing Multi-Site} air quality data set contains 6 main air pollutants and 6 relevant meteorological variables at multiple sites in Beijing. Each variable is measured hourly from March 1st, 2013 to February 28th, 2017. The goal is to predict the PM2.5 variable which is a proxy of air quality measure. Concept drifts are present in this data set due to seasonal weather changes.

\section{Experimental Design}
\label{design}
\subsection{Performance Metrics}
In the case of synthetic data sets we know exactly where the drifts occurred so we can compare the detector's outputs to them. For the synthetic data sets, we use the following metrics: detection accuracy, True positive, false positive (or false alarms) and the mean time to detection (MTD) . 
Contrary to synthetic data sets, real-world data sets don't have specified concept drift points. It is therefore very difficult to evaluate concept drift detection algorithms on them. In this case, we can't use metrics like accuracy or true positive rate to compare the detectors. To compare ADDM to the state-of-the-art algorithms on real-world data sets, we use the mean squared error (MSE) loss of the learning model in case of regression task and the cross entropy loss for classification tasks. For each data set,  if a drift is detected, a new model is learned from scratch and evaluated on a subset of the most recent data. The final performance of the detector is computed by averaging its losses on all the detected regions. 
For each detection algorithm, we also take in account the number of detections because in real-world applications a detector that gives a large number of alarms is not optimal.

\subsection{Hyper-parameters Optimization}
In this article, we have used deep learning based models as the backbone of the prediction step of the ADDM method, but any type of machine learning model can be used (logistic regression, random forest, etc.). For the synthetic data sets, our used simple multi-layer perceptron (MLP) neural networks with two hidden layers. In case of real-world data sets, we used long short-term memory (LSTM) neural networks architecture (see Table~\ref{params}). 
Each model is trained and validated on a subset of the data set where there is no concept drift. The learned model is then used to predict values for new incoming data where concept drift is suspected to be present. We then compute the error rate  of the model on new data and try to find if there are significant changes at some timestamps.
When learning with artificial neural networks, instead of monitoring the error rate which requires total access to the true labels, we can use the model's uncertainty to detect concept drift as done by Baeir et~al. \cite{baier2021detecting}. Using Monte Carlo Dropout, we can compute and monitor the model's uncertainty and detect concept drifts.

The state-of-the-art drift detection methods and algorithms ( DDM, ADWIN, PageHinkley, HDDM,KSWIN) used in this study have some hyper-parameters that should be well chosen carefully so the algorithm can adjust to the data set. For each method/algorithm, we determined the optimal hyper-parameters by using a subset of the data set that we called \textbf{experimental set}. For each synthetic data set, we took a subset containing one concept drift as the experimental set. Each algorithm is executed on the experimental sets to find the best hyper-parameters. These hyper-parameters are then used in the final experiments.
We compared these state-of-the-art methods to our ADDM drift detection method. As described in section~\ref{setarSec}, the \textbf{SETAR} model requires the user to set some hyper-parameters. The main parameters are the time delay for the threshold variable $d$, the auto regressive level $p$. In our study the parameters values are: $d=2$ and $p=5$. We aim to evaluate and give  a fair comparison among the detectors concerning the performances of real concept drift detection. The optimal hyper-parameters are listed in table~\ref{params}.

\begin{table}[htbp]
\caption{Description and Neural network architecture of each data set. Drifts refers to the number of drifts.The remaining columns show, the optimal hyper-parameter according to the detector.}
\label{params}
\begin{center}
\resizebox{\columnwidth}{!}{
\begin{tabular}{|l|c|c|c|c|c|c|c|c|c|c|}
\hline
{\textbf{Data set}}&\textbf{Samples} &\textbf{Features}&\textbf{Target} &\textbf{Drifts}& \textbf{architecture} & \textbf{Hyper-parameters} &  \textbf{Epochs} & \textbf{ADWIN} & \textbf{Page-Hinkley} & \textbf{KSWIN} \\
 \hline
Friedman &20000 &3& continuous & 3 &MLP & (30,15,1) & 50 &$1\times10^{-4}$&  $1\times10^{-6}$ &0.001 \\
Friedman no return &20000 &3& continuous &6&MLP & (30,15,1) & 50& $1\times10^{-6}$ & $1\times10^{-6}$  &0.0034  \\
Brieman 2d planes &20000&11& continuous &6 &MLP & (30,15,1) & 50 & $1\times10^{-3}$  & $1\times10^{-6}$&0.0032 \\
Agrawal 32 &20000&3& 2 classes &1 &MLP &   (20,10,1) &  50& $5\times10^{-6}$  & 50&0.0034 \\
Agrawal 3213 &20000 &3&  2 classes & 4 &MLP &   (20,10,1) &  50 & $4.2\times10^{-5}$ & 50 &0.0029\\
Mixed  & 20000&5& continuous &3 &MLP &   (20,10,1) &  50& 0.0441  & $1\times10^{-6}$& 0.0059\\
Panama electricity & 48048 & 17& continuous &- &LSTM &    (120,60,30) & 50& $1\times10^{-5}$ & $1\times10^{-10}$ &$1\times10^{-10} $\\
Italian air quality &  8991 & 14& continuous &- &LSTM &    (64,32,15) & 50 & $3.41\times10^{-4} $ & $1\times10^{-10}$ &$1\times10^{-10}$ \\
NSW electricity & 45312 & 9& 2 classes &- &LSTM &    (120,60,30)& 50&  $1\times 10^{-5}$ & $1\times10^{-10}$ &$1\times10^{-10}$\\
Gas sensor drift & 13910 & 129& 6 classes &- & MLP &    (120,120,6)  & 50  &  $4\times10^{-15}$ &  0.05 & $1\times10^{-10}$  \\
Beijing air quality & 420768 & 18& continuous &- &LSTM &    (120,60,30) & 50  & $6\times10^{-5}$ &  $1\times10^{-10}$ &$1\times10^{-10}$ \\
\hline
\end{tabular} 
}
\end{center}
\end{table}

\section{Results and Discussions}
In this section, we present the results of the state-of-the-art drift detection methods and ADDM method on the experimental data sets. For each data set, we compared ADDM to seven state-of-the-art concept drift detection methods.

\subsection{Results on Synthetic Data sets}
\begin{figure}[htbp]
     \centering
     \begin{subfigure}[b]{0.49\textwidth}
         \centering
         \includegraphics[width=\textwidth]{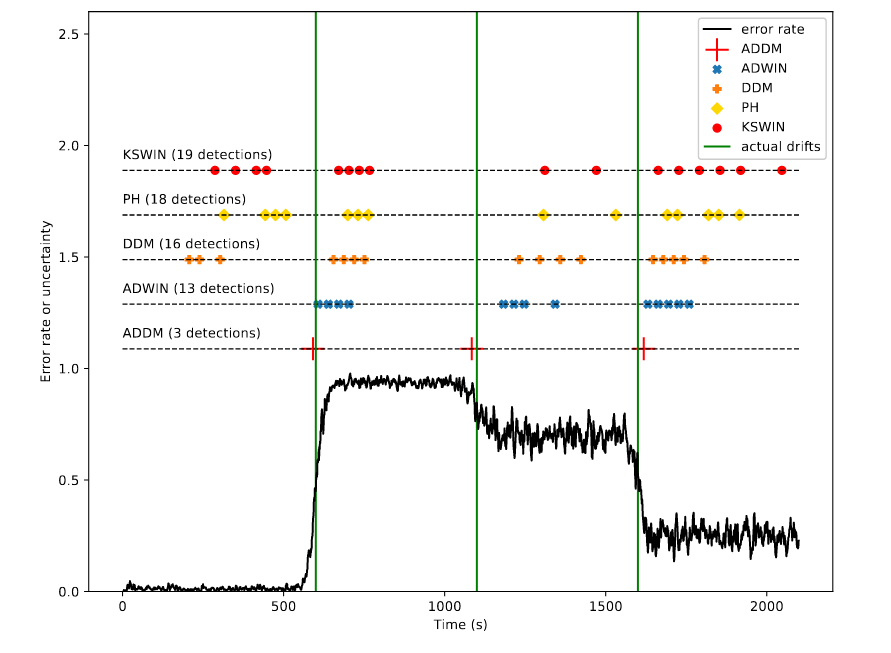}
       \caption{Detected drifts on the \textbf{Mixed} data set. The vertical lines in green are the actual drifts points. The horizontal lines show the detected drifts by each detector. The horizontal line with red plus (+) markers shows our method's detections.}
         \label{mixed}
     \end{subfigure}
     \hspace{0.5em}%
     \begin{subfigure}[b]{0.49\textwidth}
         \centering
        \includegraphics[width=\textwidth]{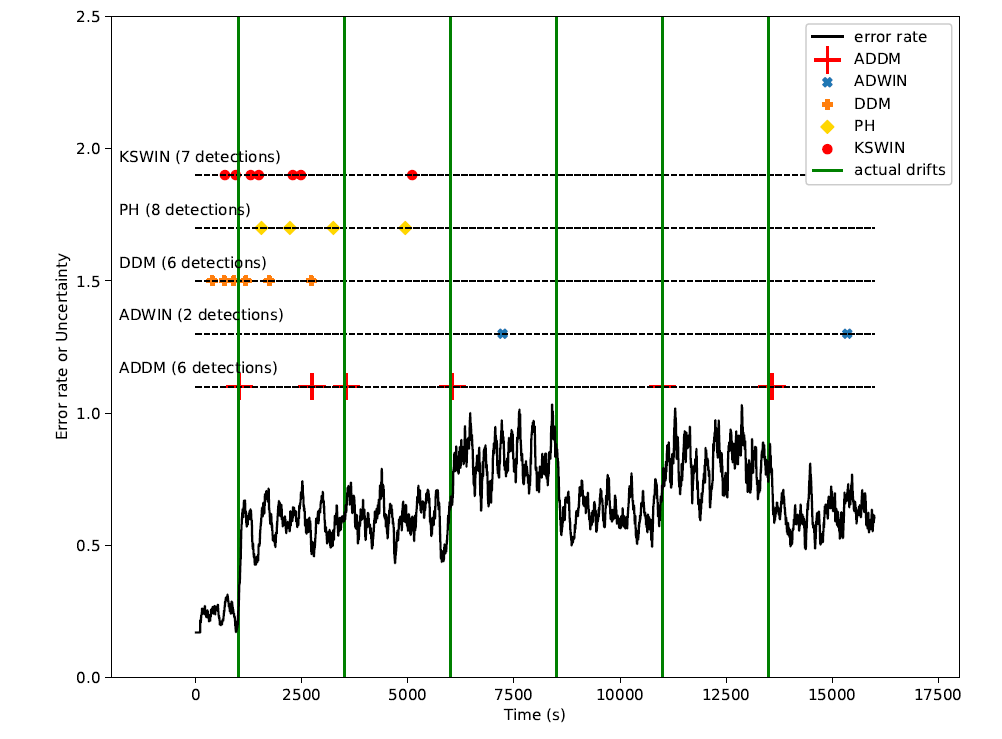}
         \caption{Detected drifts on the \textbf{Brieman 2d planes} data set. The vertical lines in green are the actual drifts points. The horizontal lines show the detected drifts by each detector. The horizontal line with red plus (+) markers shows our method's detections. }
         \label{briedman}
     \end{subfigure}
     \caption{Detector performance on synthetic data sets}
     \label{fig:w1}

\end{figure}

Table \ref{results} shows the experimental results of the drift detection algorithms on the synthetic data sets. 
Recall that in case of synthetic data sets we know the exact drift points so we can compare them to the detector's outputs. These results show that ADDM outperforms all other methods in terms of true positives (TP) and false alarms (FA). It has a very low false alarm rate. Despite the parameter optimization , note that the state-of-the-art methods detect a very large number of drifts. This illustrates these algorithms' problem of high reactivity leading to a large number of false positive drift detection \cite{baier2021detecting}. Contrary to state-of-the-art methods, ADDM is less sensitive to small variation and only detects statistically significant drifts. This is very important in real-world application because retraining a machine model is time-consuming and resource-intensive. 
Fig.~\ref{mixed} and Fig.~\ref{briedman} show the results of all the algorithms applied respectively to the \textbf{Mixed} and the \textbf{ Brieman 2d planes} data sets. On these data sets, ADDM accurately detects all the drifts unlike the other algorithms. The \textbf{Mixed} data set contains three concept drift points and ADDM  is the only method capable of accurately detecting all the drift points (horizontal line with red plus (+) markers in Fig.~\ref{mixed}). The \textbf{Brieman} data set contains six (6) drift points, ADDM accurately detected five of them (horizontal line with red plus (+) markers in Fig.~\ref{briedman}) when none of the other methods detects more than one drift. 
\begin{table}[htbp]
\caption{ Detector performance on synthetic data sets. The columns show the false alarms (FA), true positives (TP) and Mean Time to Detection (MTD) of the detector. The MTD is expressed in seconds}
\label{results}
\begin{center}
\resizebox{0.8\columnwidth}{!}{%
\begin{tabular}{|lcc|c|c|c|c|c|c|}
\hline
\multirow{1}{*}{\textbf{Method}} & &\textbf{Metric}& \textbf{Agrawal 32} &  \textbf{Agrawal 3213} & \textbf{Mixed} &  \textbf{\textit{Friedman}} &  \textbf{friedman no return} &  \textbf{Brieman 2d planes}  \\
\hline
\multicolumn{2}{|l}{\multirow{3}{*}{\textbf{ADDM}}} & \textbf{\textit{TP}}  & 1 & 2 &        3 &  3 & 5 & 5 \\
\multicolumn{2}{|c}{}&  \textbf{\textit{FA}}  & 0 & 2 &0 &  0 & 1 & 1 \\
\multicolumn{2}{|c}{}&  \textbf{\textit{MTD}}  &  1.5 &  3.657 &  0.368 &  21.823 &  11.912 &  20.711 \\ 
\hline
\multicolumn{2}{|l}{\multirow{3}{*}{ADWIN}} &TP  & 0 & 4 & 2 &  1 & 0 & 0 \\
\multicolumn{2}{|c}{}&  FA  & 7 & 15 &       11 &  7 & 10 & 2 \\
\multicolumn{2}{|c}{}&  MTD  &  0.342 &  0.657 &  0.128 &  1.182 &  0.935 &  1.289 \\
\hline
\multicolumn{2}{|l}{\multirow{3}{*}{DDM}} & TP    & 0 & 2 &        1 &  0 & 1 &   1 \\
\multicolumn{2}{|c}{}&  FA    & 17 & 21 &       15 &  6 & 16 & 5 \\
\multicolumn{2}{|c}{}&  MTD&  0.372 &  0.612 &  0.126 &  1.435 &  0.916 &  1.348 \\
\hline
\multicolumn{2}{|l}{\multirow{3}{*}{KSWIN}} &TP  & 0 & 0 &        0 &  0 & 1 & 1 \\
\multicolumn{2}{|c}{}&  FA  & 18 & 21 &       19 &  4 & 9 & 6 \\
\multicolumn{2}{|c}{}&  MTD&  0.398 &  0.690 &  0.178 &  1.383 &  1.103 &  1.287 \\
\hline
\multicolumn{2}{|l}{\multirow{3}{*}{PH}}& TP     & 0 & 1 &        1 &  0 & 0 & 1 \\
\multicolumn{2}{|c}{}&  FA     & 15 & 19 &       17 &  3 & 11 & 7 \\
\multicolumn{2}{|c}{}&  MTD&  0.411 &  0.613 &  0.138 &  1.229 &  0.980 &  1.360 \\
\hline
\multicolumn{2}{|l}{\multirow{3}{*}{EDDM}}& TP   & 0 & 5 &        3 &  0 & 1 & 3 \\
\multicolumn{2}{|c}{}&  FA   & 14 & 22 &       17 &  4 & 17 & 3 \\
\multicolumn{2}{|c}{}&  MTD&  0.355 &  0.611 &  0.134 &  1.336 &  1.044 &  1.247 \\
\hline
\multicolumn{2}{|l}{\multirow{3}{*}{HDDM\_A}}& TP  & 0 & 2 &3 &  0 & 2 & 3 \\
\multicolumn{2}{|c}{}&  FA  & 17 & 23 &       15 &  5 & 17 & 3 \\
\multicolumn{2}{|c}{}&  MTD &  0.384 &  0.819 &  0.140 &  1.373 &  1.012 &  1.388 \\
\hline
\multicolumn{2}{|l}{\multirow{3}{*}{HDDM\_W}}& TP  & 0 & 1 &        3 &  0 & 0 & 3 \\
\multicolumn{2}{|c}{}&   FA  & 17 & 19 &       18 &  6 & 16 & 3 \\
\multicolumn{2}{|c}{}&  MTD&  0.403 &  0.630 &  0.133 &  1.343 &  1.236 &  1.324 \\
\hline
\end{tabular}
}
\end{center}
\end{table}

\subsection{Results on Real-world Data sets}
Table~\ref{real_world_results} shows the results of the drift detection algorithms on real-world data sets.
Recall that when comparing ADDM to state-of-the-art algorithms on real-world data sets, we use the mean squared error loss of the learning model in case of regression task and the cross entropy loss for classification tasks.
In order to have a better view of the detector's performances, we also listed the number of times the model was retrained ($nb\_train$). The number of retraining is equal to the number of detected drifts and gives us an idea of how sensitive the detector is. As expected, the state-of-the-art detectors lead to a large number of retraining. By combining the loss and the number of retraining,  in almost all cases ADDM outperforms the state-of-the-art detectors. 
In the rare cases where other methods have outperformed ADDM, the improvement is very small and the retraining is at least two times that of ADDM. For example, on the Gas sensor drift data set, the KSWIN algorithm retrained the model 42 times (loss=1.69) while ADDM retrained the model only 7 times (loss=1.89).
 
\begin{table}[htbp]
\caption{Performance of detectors on real-world data sets. The model's loss (the lower the best) and the  number of retrainings $nb\_train$ (the lower the less computationally expensive).}
\begin{center}
\resizebox{\columnwidth}{!}{%
\begin{tabular}{|llc|c|c|c|c|c|}
\hline
\multirow{1}{*}{\textbf{Method}} & &\textbf{Metric}&\textbf{Panama electricity} & \textbf{Italian air quality }  &\textbf{Beijing air quality} &\textbf{Gas sensor drift}  & \textbf{NSW data set} \\
\hline
\multicolumn{2}{|l}{\multirow{2}{*}{\textbf{ADDM}}} & \textbf{\textit{loss}}  &  \textbf{0.020} &    \textbf{0.007} &    \textbf{0.006} &   \textbf{1.89} &   \textbf{0.290} \\
\multicolumn{2}{|l}{}& \textbf{\textit{nb\_retrain}} &   \textbf{7} &  \textbf{5} &  \textbf{5} &  \textbf{7} &  \textbf{5} \\
\hline
\multicolumn{2}{|l}{\multirow{2}{*}{ADWIN}} & loss &   0.032 & 0.02 &    0.016 &   1.727 &   0.234 \\
\multicolumn{2}{|l}{}& nb\_retrain &   7 &  0 &  195 &  21 &  12 \\
\hline

\multicolumn{2}{|l}{\multirow{2}{*}{DDM}} &loss   &   0.030 &  0.014 &    0.014 &   1.714 &   0.253 \\
\multicolumn{2}{|l}{}& nb\_retrain  &  18 &  4 &  145 &  41 &  17 \\
\hline
\multicolumn{2}{|l}{\multirow{2}{*}{KSWIN}} &loss &   0.034 &  0.011 &    0.016 &   1.69 &   0.255 \\
\multicolumn{2}{|l}{}& nb\_retrain  &  10 &  6 &  132 &  42 &  10 \\
\hline
\multicolumn{2}{|l}{\multirow{2}{*}{PH}} &loss    &   0.044 &  0.016 &    0.016 &   1.724 &   0.239 \\
\multicolumn{2}{|l}{}& nb\_retrain     &  12 &  7 &  119 &  37 &  11 \\
\hline
\multicolumn{2}{|l}{\multirow{2}{*}{EDDM}} & loss        &   0.026 &  0.008 & 0.004  &   1.748& 0.276    \\
\multicolumn{2}{|l}{}& nb\_retrain   &  10 &  3 &  78 &  33&8 \\
\hline
\multicolumn{2}{|l}{\multirow{2}{*}{HDDM\_A}} & loss       &   0.058 &  0.007 & 0.003  & 1.694& 0.260\\
\multicolumn{2}{|l}{}& nb\_retrain &   5 &  4 &  69 & 23&  9 \\
\hline
\multicolumn{2}{|l}{\multirow{2}{*}{HDDM\_W}} & loss       &   0.059 &  0.009 & 0.004  &    1.696&  0.288   \\
\multicolumn{2}{|l}{}& nb\_retrain  &   5 &  3 &  58 &   27&8  \\
\hline
\end{tabular} }
\end{center}
\label{real_world_results}
\end{table}

\section*{ Conclusion}
Detecting concept drift is important in real-world applications as it leads to a decrease in machine learning models performance. The traditional concept drift detection methods are very sensitive to changes and leads to a large number of false alarms. These methods also often require full access to the true labels. In this paper, we propose a method that combines a machine learning algorithm with autoregressive time series models to detect concept drift in stream data. The main idea is to consider the error rate of a machine learning model as a time series and model them with an autoregressive time series model. We compared ADDM to seven (7) state-of-the-art concept drift detection algorithms on six (6) synthetic data sets and five (5) real-world data sets. The results show that it outperforms all the state-of-the-art algorithms in terms of accuracy and has a very low false alarm rate. In addition to the drift detection method, we proposed a new method of concept drift adaptation based on the severity of the drift. The main idea is to aggregate the old and new model using an estimate of the dissimilarity between the old concept and the new one as weights. The higher is the severity, the less relevant is the old model. 
In future works we aim to use auto-regressive models to detect concept drifts using directly the input data instead of the error rate or the model's uncertainty. This can be very useful in cases where the true labels are not available.
\bibliographystyle{unsrt} 
\bibliography{references}

\begin{thebibliography}{10}

\bibitem{baier2021detecting}
Lucas Baier, Tim Schl{\"o}r, Jakob Sch{\"o}ffer, and Niklas K{\"u}hl.
\newblock Detecting concept drift with neural network model uncertainty.
\newblock {\em arXiv preprint arXiv:2107.01873}, 2021.

\bibitem{kadam2019survey}
Shweta Kadam.
\newblock A survey on classification of concept drift with stream data.
\newblock 2019.

\bibitem{tong2009threshold}
Howell Tong and Keng~S Lim.
\newblock Threshold autoregression, limit cycles and cyclical data.
\newblock In {\em Exploration Of A Nonlinear World: An Appreciation of Howell
  Tong's Contributions to Statistics}, pages 9--56. World Scientific, 2009.

\bibitem{tsay1989testing}
Ruey~S Tsay.
\newblock Testing and modeling threshold autoregressive processes.
\newblock {\em Journal of the American statistical association},
  84(405):231--240, 1989.

\bibitem{gonzalo2005subsampling}
Jesus Gonzalo and Michael Wolf.
\newblock Subsampling inference in threshold autoregressive models.
\newblock {\em Journal of Econometrics}, 127(2):201--224, 2005.

\bibitem{lu2018learning}
Jie Lu, Anjin Liu, Fan Dong, Feng Gu, Joao Gama, and Guangquan Zhang.
\newblock Learning under concept drift: A review.
\newblock {\em IEEE Transactions on Knowledge and Data Engineering},
  31(12):2346--2363, 2018.

\bibitem{bifet2007learning}
Albert Bifet and Ricard Gavalda.
\newblock Learning from time-changing data with adaptive windowing.
\newblock In {\em Proceedings of the 2007 SIAM international conference on data
  mining}, pages 443--448. SIAM, 2007.

\bibitem{gama2004learning}
Joao Gama, Pedro Medas, Gladys Castillo, and Pedro Rodrigues.
\newblock Learning with drift detection.
\newblock In {\em Brazilian symposium on artificial intelligence}, pages
  286--295. Springer, 2004.

\bibitem{page1954continuous}
Ewan~S Page.
\newblock Continuous inspection schemes.
\newblock {\em Biometrika}, 41(1/2):100--115, 1954.

\bibitem{raab2020reactive}
Christoph Raab, Moritz Heusinger, and Frank-Michael Schleif.
\newblock Reactive soft prototype computing for concept drift streams.
\newblock {\em Neurocomputing}, 416:340--351, 2020.

\bibitem{gama2006learning}
Joao Gama and Gladys Castillo.
\newblock Learning with local drift detection.
\newblock In {\em International conference on advanced data mining and
  applications}, pages 42--55. Springer, 2006.

\bibitem{baena2006early}
Manuel Baena-Garc{\i}a, Jos{\'e} del Campo-{\'A}vila, Ra{\'u}l Fidalgo, Albert
  Bifet, R~Gavalda, and Rafael Morales-Bueno.
\newblock Early drift detection method.
\newblock In {\em Fourth international workshop on knowledge discovery from
  data streams}, volume~6, pages 77--86, 2006.

\bibitem{frias2014online}
Isvani Frias-Blanco, Jos{\'e} del Campo-{\'A}vila, Gonzalo Ramos-Jimenez,
  Rafael Morales-Bueno, Agustin Ortiz-Diaz, and Yaile Caballero-Mota.
\newblock Online and non-parametric drift detection methods based on
  hoeffding’s bounds.
\newblock {\em IEEE Transactions on Knowledge and Data Engineering},
  27(3):810--823, 2014.

\bibitem{xu2017dynamic}
Shuliang Xu and Junhong Wang.
\newblock Dynamic extreme learning machine for data stream classification.
\newblock {\em Neurocomputing}, 238:433--449, 2017.

\bibitem{yan2020accurate}
Myuu Myuu~Wai Yan.
\newblock Accurate detecting concept drift in evolving data streams.
\newblock {\em ICT Express}, 6(4):332--338, 2020.

\bibitem{gama2013evaluating}
Joao Gama, Raquel Sebastiao, and Pedro~Pereira Rodrigues.
\newblock On evaluating stream learning algorithms.
\newblock {\em Machine learning}, 90(3):317--346, 2013.

\bibitem{6706768}
Shuo Wang, Leandro~L. Minku, Davide Ghezzi, Daniele Caltabiano, Peter Tino, and
  Xin Yao.
\newblock Concept drift detection for online class imbalance learning.
\newblock In {\em The 2013 International Joint Conference on Neural Networks
  (IJCNN)}, pages 1--10, 2013.

\bibitem{9679880}
Salvatore Greco and Tania Cerquitelli.
\newblock Drift lens: Real-time unsupervised concept drift detection by
  evaluating per-label embedding distributions.
\newblock In {\em 2021 International Conference on Data Mining Workshops
  (ICDMW)}, pages 341--349, 2021.

\bibitem{hansen1996inference}
Bruce~E Hansen.
\newblock Inference in tar models.
\newblock {\em Unpublished working paper. Chestnut Hill, MA: Boston College
  Department of Economics (December)}, 1996.

\bibitem{skmultiflow}
Jacob Montiel, Jesse Read, Albert Bifet, and Talel Abdessalem.
\newblock Scikit-multiflow: A multi-output streaming framework.
\newblock {\em Journal of Machine Learning Research}, 19(72):1--5, 2018.

\bibitem{friedman1991multivariate}
Jerome~H Friedman.
\newblock Multivariate adaptive regression splines.
\newblock {\em The annals of statistics}, pages 1--67, 1991.

\bibitem{agrawal1993database}
Rakesh Agrawal, Tomasz Imielinski, and Arun Swami.
\newblock Database mining: A performance perspective.
\newblock {\em IEEE transactions on knowledge and data engineering},
  5(6):914--925, 1993.

\bibitem{Dua2019}
Dheeru Dua and Casey Graff.
\newblock {UCI} machine learning repository, 2017.

\bibitem{aguilar2021short}
Ernesto Aguilar~Madrid and Nuno Antonio.
\newblock Short-term electricity load forecasting with machine learning.
\newblock {\em Information}, 12(2):50, 2021.

\bibitem{de2008field}
Saverio De~Vito, Ettore Massera, Marco Piga, Luca Martinotto, and Girolamo
  Di~Francia.
\newblock On field calibration of an electronic nose for benzene estimation in
  an urban pollution monitoring scenario.
\newblock {\em Sensors and Actuators B: Chemical}, 129(2):750--757, 2008.

\bibitem{aqicn}
About the world air quality index project.

\bibitem{vergara2012chemical}
Alexander Vergara, Shankar Vembu, Tuba Ayhan, Margaret~A Ryan, Margie~L Homer,
  and Ram{\'o}n Huerta.
\newblock Chemical gas sensor drift compensation using classifier ensembles.
\newblock {\em Sensors and Actuators B: Chemical}, 166:320--329, 2012.

\end{thebibliography}
\end{document}